\documentclass[journal]{IEEEtran}
\ifCLASSINFOpdf
\else
\fi

\usepackage{footmisc}
\usepackage{subfigure}
\usepackage{url}
\usepackage{multirow}

\usepackage[noadjust]{cite}
\usepackage{amsmath,amsthm}
\usepackage{amssymb,amsfonts}
\usepackage{booktabs}
\usepackage{color}
\usepackage{ccaption}
\usepackage{booktabs}
\usepackage{float}
\usepackage{fancyhdr}
\usepackage{caption}
\usepackage{xcolor,stfloats}
\usepackage{graphicx}
\usepackage{algpseudocode}
\usepackage[ruled,linesnumbered]{algorithm2e}

\usepackage[utf8]{inputenc}
\usepackage{footmisc}
\usepackage{subfigure}

\begin{document}
%
\title{Searching Similarity Measure for  Binarized \\Neural Networks}
%
%
%


\author{Yanfei~Li,
        Ang~Li
        and~Huimin~Yu
\thanks{Y. Li and H. Yu are with Zhejiang University, Zhejiang,
China.}
\thanks{A. Li is with Pacific Northwest National Laboratory, Richland, WA.}
}

\maketitle

\begin{abstract}

Being a promising model to be deployed in resource-limited devices, Binarized Neural Networks (BNNs) have drawn extensive attention from both academic and industry. However, comparing to the full-precision deep neural networks (DNNs), BNNs suffer from non-trivial accuracy degradation,  limiting its applicability in various domains. This is partially because existing network components, such as the similarity measure, are specially designed for DNNs, and might be sub-optimal for BNNs.

In this work, we focus on the key component of BNNs -- the similarity measure, which quantifies the distance between input feature maps and filters, and propose an automatic searching method, based on genetic algorithm, for BNN-tailored similarity measure. Evaluation results on Cifar10 and Cifar100 using ResNet, NIN and VGG show that most of the identified similarty measure can achieve considerable accuracy improvement (up to 3.39\%) over the commonly-used cross-correlation approach.

\end{abstract}


%
\IEEEpeerreviewmaketitle

\section{Introduction}

\IEEEPARstart{D}{eep} neural networks (DNNs) have been already widely adopted and proved to be promising in various domains, such as computer vision \cite{assael2016lipnet,cao2017realtime}, language processing \cite{xiong2016achieving,ping2017deep}, speech recognition \cite{severyn2015twitter, hermann2015teaching}, bioinformatics \cite{ma2015deep,leung2014deep}, etc. However, their massive computing demand hampers their applicability in resources-limited terminal devices, such as smart phone, camera, robots, wearable devices, etc. To reduce such computation cost, existing effort aim at simplifying the computation and proposing more hardware-friendly models. Binarized neural networks (BNNs) \cite{courbariaux2016binarized, hubara2016binarized} serve as one of such effort and draw much attention recently.

BNNs binarize the floating-point inputs and weights of DNNs into a binary value: +1 and -1, which naturally maps to 1 and 0 in digital logic, thus is extremely hardware-friendly. Binarization is the extreme of the quantization approach and exhibits several nice features: (i) \emph{Memory efficiency}: as both inputs and weights are represented by bits, comparing to single-precision floating format, BNN can theoretically reduce the capacity and bandwidth demand by 32x for the entire memory hierarchy; (ii) \emph{Execution efficiency}: the expensive floating-point dot-product (also known as multiply-accumulate) operations can be replaced by bit-wise exclusive-nor (XNOR) and population-count operations (POPC), improving execution efficiency by more than 10$\times$ \cite{li2019bstc}. (iii) \emph{Low latency}: Given the computation and memory benefit, the raw latency of single image inference can be reduced drastically by 58x on CPUs \cite{courbariaux2016binarized}, and more than 1000x on GPU \cite{li2019bstc} and other systems \cite{kung2018efficient}, allowing read-time prediction for latency-critical applications; (iv) \emph{Power efficiency:} due to simplified hardware and less memory demand, BNNs can run on extremely energy-constraint edge devices; Last but not the least, (v) \emph{Robustness}: it has been recently shown that, due to the discrete nature through binarization, compared to normal DNNs, BNNs show improved robustness against noise and poison attacks \cite{lin2019defensive}. Because of these features, BNNs have been showcased on a variety of practical edge applications, such as COVID-19 face-cover detection \cite{fasfous2021binarycop}, auto-driving \cite{chen2020gpu}, smart agriculture \cite{huang2021fpga}, image enhancement \cite{ma2019efficient}, 3D objection detection \cite{ma2018binary}, etc. 

Owing these nice features, current BNNs suffer from non-trivial accuracy degradation. This is largely due to information loss and gradient approximation for the binarization process, as well as the mismatch between BNN semantics and the present network structures that are traditionally designed mainly for DNNs. To close this accuracy gap, various effort have been posed, as briefly summarized in Section~II. In this paper, we focus on the similar measure, which quantifies the distance between the binary input feature maps and filters for binary deep convolutional neural networks.

Deep convolutional neural networks (CNNs) have shown exemplary performance in many computer vision applications, such as image classification \cite{krizhevsky2012imagenet, he2016deep}, object detection \cite{ren2015faster}, visual tracking \cite{nam2016modeling},
semantic segmentation \cite{long2015fully}, action recognition \cite{lan2017deep}, etc.
The success of CNNs largely attributes to its convolution operations, which rely on a similarity measure, such as cross-correlation, to estimate the distance between two features. The convolution thus can be seen as matching the predefined feature templates (i.e., the filters) to the input patch, and obtain the correspondence degree. A low value thus implies the irrelevance of the input patch to the feature template, and will likely to be filtered out in subsequent layers. The learning process is to figure out the best feature templates per layer through back-propagation. As can be seen, the similarity measure is critical for the convolution operation. 

For DNNs, cross-correlation is the standard similarly measure, and few attempt have been posed for replacing it due to its excellent performance. However, for BNNs, both the activation and weights are binary vectors. Therefore, the full-precision tailored cross-correlation metric might not be the silver bullet for BNNs. 

As binary vectors are among the most commonly used feature representations, many similarity measure approaches have been proposed so far, for distinct domain purposes \cite{consonni2012new,todeschini2012similarity,choi2010survey}. For example, the Jaccard similarity measure \cite{jaccard1901etude} are widely adopted for ecology and biology tasks. Since BNN is a new domain, it is unknown whether any of these existing binary similarity measures, or their combination, can be more optimal than the default cross-correlation baseline. 

In this work, we attempt to answer this question through automatic searching. Since most existing binary similarity measures are crafted based on human expertise towards distinct purposes, it is non-feasible nor cost-effective to traverse the large design space by thoroughly try-out each of them and their combinations. An automatic approach thus exhibits great advantage, as already showcased in discovering novel activation functions \cite{ramachandran2017searching}, network architectures \cite{liu2018darts}, and normalization layers \cite{liu2020evolving}.

In this paper, we use the genetic algorithm (GA) based searching algorithm to identify appropriate binary similarity measure for BNNs. We represent the the similarity measure with a fixed graph, each node indicates a operator, which is selected from a list of candidates. The optimal combination of different operators is the promising similarity measure, which is auto-searched by GA. The top 10 promising similarity measures are listed and evaluated on Cifar10 and Cifar100 with several network models. Evaluation results show that the newly obtained similarity measures can achieve better accuracy, with a maximum accuracy improvement of 3.39\% on Cifar100. 
Our contributions in this paper are:

\begin{itemize}
\item we propose to discover more appropriate similarity measure for BNNs to achieve better accuracy.
\item Through GA-based searching, we have identified promising similarity measures, which show enhanced accuracy of BNNs.
\end{itemize}

The remainder of this paper is organized as follows.
Section~II briefly introduces related work. Section~III presents the way of searching similarity measure by genetic algorithm. Experiments are shown in Section~IV, and conclusion are drawn in Section~V.

\section{related work}
\subsection{BNN}
BNNs \cite{courbariaux2016binarized, hubara2016binarized} use binary values for activations and weights in the neural network, while DNNs use full-precision values.  Thanks to the binary values, most computations in BNNs can be executed by bitwise operations, which can largely save computational resources, memory demands and powers. However BNNs also suffer the severe accuracy drop compared to DNNs, which is mainly due to the information loss and gradient approximation bring by binarization process. 

Since BNNs were proposed, there are continuous efforts focusing on improving the accuracy. Lots of strategies have been proposed and have shown successful improvement:
\begin{itemize}
    \item Minimize the quantization error. 
    XNOR-Net \cite{rastegari2016xnor} introduced scaling factor during the binarization to less quantization error. Similar idea was proposed in BWNH \cite{hu2018hashing} that considered the quantization as hash map with scaling factors. XNOR-Net++ \cite{bulat2019xnor} used a learnable scaling factor which can be learned during backward propogation. HORQ \cite{li2017performance} adopted a recursive approximation based on the quantized residual. In ABC-Net \cite{lin2017towards} the full-precision weights and activations were approximately as a linear combination of multiple binary weight matrices with learned terms.

    \item Reduce the gradient error. 
    To deal with the gradients for the non-differential binarization function, straight-through estimator (STE) technique was adopted in BNNs to estimate the gradients in backward propagation, which leads to under-optimized  binary networks with performance degradation. Bi-Real \cite{liu2018bi} proposed using a second order function ApproxSign to replace sign during back-propagation. BNN+ \cite{darabi2018bnn+} presented a variation of the derivative of the Swish-like activation for an effective back-propagation function. Lahoud et al. \cite{lahoud2019self} proposed a soft quantization method, which adopted the scaled hyperbolic tangent function to progressively reshape the activations toward binary values.
    Hou et al. \cite{hou2016loss} discussed loss-aware binarization, showcasing a proximal Newton algorithm with diagonal Hessian approximation that could directly minimize the loss with respect to the binary weights. IR-Net \cite{qin2020forward} introduced a self-adaptive error decay estimator in training, which retain both forward and backward information.
    
    \item Design BNN-specific network structure. Most network models used in BNNs were classic network architecture that designed for DNNs,  researchers have started to design BNN-oriented network structures. Bi-Real \cite{liu2018bi} proposed to adopt more shortcuts for reusing information to maintain rich information flow in the network. BinaryDenseNet \cite{bethge2019binarydensenet} concatenated the generated features and used full-precision convolution in downsampling layers to increase the information flow. 
    MeliusNet \cite{bethge2021meliusnet} consisted of alternating DenseBlock and Improvement Block to increase the feature capacity and quality. 
    BENN \cite{zhu2019binary}  aggregate multiple BNNs by boosting or bagging to improve the accuracy.
    
    \item Tricks for training. 
    Tang et al. \cite{tang2017train} presented special regularization for BNNs to encourage the latent floating-point variables approaching +1 and -1 during the training.
    Alizadeh et al. \cite{alizadeh2018empirical} empirically studied various optimizer and found that
    adapting the learning rate using second-moment methods was crucial for the successful use of STE. Binary Optimizer \cite{helwegen2019latent} introduced a BNN-specific optimizer by viewing the latent weights as inertia.
    INQ \cite{zhou2017incremental} guided the training of BNNs through a full-precision teacher network by adding a regularization term in loss function. 
\end{itemize}

More works about BNNs could be found in the two surveys \cite{simons2019review, qin2020binary}. In this paper, we propose to discover more appropriate similarity measure for BNNs to improve its accuracy.

\subsection{Genetic Algorithm}
Meta-heuristic algorithms are used to effectively solve complex problems, such as travelling salesman problem, scheduling problem, space allocation problem, clustering problem, etc. The well-known meta-heuristics are tabu search, simulated annealing, genetic algorithm, ant colony optimization, particle swarm optimization, etc. Among those, genetic algorithm (GA) is one of the population-based algorithms, and have shown its suitability to solve combinatorial problem \cite{ouarda2014comparison}.

GA is proposed by John Holland \cite{holland1992genetic} and enhanced by David Goldberg \cite{goldberg1988genetic}. The main idea of GA is to mimic the natural selection and the survival of the fittest. GA can deal with various types of optimization, whether the objective function is stationary or change with time, linear or nonlinear, continuous or discontinuous, or with random noise. There are many variants of genetic algorithms, which have been applied to a wide range of fields. For example computer science, bioinformatics, earth sciences, finance and economics.

In GA, the candidate solution is represented as chromosome, which is an individual in the population. Fitness function that related to the objective function is used to evaluate the degree of goodness of candidate solutions. GA starts with a initialization population with randomly generated individuals. Individuals in the population are evaluated for fitness values and ranked from best to worst based on fitness values. Then the evolution process begins. In each generation, offspring is generated by genetic operations (selection, crossover and mutation). The best individuals among the population alongside the new offspring form the new population, which will be used for the next generation. The evolution process repeat until the population has converged (not producing any offspring that are significantly different from the previous generation) or the maximum number of generations is reached. In GA, the population can explore the search space in many directions simultaneously, which is ideal to parallelize the implementation.

GA is an appealing approach to obtain high-quality solutions for an optimization problem thanks to its ability to deliver a good solution fast-enough. For GA, there is no absolute assurance to find a global optimum, but it comes with a population of solutions which are satisifying enough to the problem. Please refer to \cite{thengade2012genetic,katoch2020review} for a detailed review about GA.

\subsection{Binary Similarity Measure}
Binary feature vectors have been commonly used to represent patterns in many problems such as clustering, classification, etc. The binary variables are used to reflect either the present or absence of certain attributes. Numerous binary similarity measures have been proposed, and different similarity measures have had a meaningful performance in their respective field. Applying appropriate measure is crucial for the problem.

Assuming there are two binary vectors $x$ and $y$, whose element is either 1 or 0, indicating the corresponding attribute is presence or absence. The association coefficients are calculated as shown in Table~\ref{abcd}. There are four combinations for two binary variables: 1/1, 0/1, 1/0, 0/0. 
$a$, $b$, $c$, $d$ are the frequencies of the corresponding combinations, and $n$ is the length of both binary vectors, which is equal to $a+b+c+d$.

\begin{table}[htbp]
\centering
\caption{ Frequencies of four possible combinations}
\begin{tabular}{c| c| c| c} 
\hline
\hline
          & $x=1$ & $x=0$ & Sum\\
\hline
 $y =1$  & $a = x \cdot y$  & $b = \overline{x} \cdot y$ & $a+b$ \\
\hline
$y =0$  & $c = x \cdot \overline{y}$  & $d = \overline{x} \cdot \overline{y}$ & $c+d$ \\
\hline
Sum & $a+c$ & $b+d$ & $n = a+b+c+d$ \\
\hline
\hline
\end{tabular}
\label{abcd}
\end{table}

$a$: the number of attributes where both $x$ and $y$ share (1/1), which is "positive matches".

$b$: the number of attributes where $x$ lacks and $y$ has (0/1).

$c$: the number of attributes where $x$ has and $y$ lacks (1/0).

$d$: the number of attributes where both $x$ and $y$ lack (0/0), which is "negative matches".

$a+d$: the total number of matched attribute between $x$ and $y$, both have or lack (1/1 or 0/0).

$b+c$: the total number of mismatched attributes where $x$ has $y$ lacks or $x$ lacks $y$ has  (1/0 or 0/1).

$n=a+b+c+d$: the length of binary vectors.

Since Jaccard proposed a similarity measure to classify ecological species, numerous similarity measurement have been proposed. 
Most of binary similarity measures are functions of $a$, $b$, $c$, $d$. The ratio of matches attributes and mismatches attributes is commonly used to measure the similarity, as shown in Eq~\ref{ad_1}, $w$ is the weighted factor of mismatched attributes.  
\begin{equation}
 S=\frac{a+d}{a+d+w(b+c)}
 \label{ad_1}
\end{equation}

The measure is Sokal \& Michener if $w=1$. When $w=2$, the measure is Rogers Tanimoto, and is Gower Legendre if $w=1/2$.

Some measures exclusion of negative matches ($d$), i.e. Eq~\ref{ad_2}. When $w=1$, the measure is Jaccard. If $w=2$, the measure is Sokal \& Sneath, and  is Dice when $w=1/2$.
\begin{equation}
 S=\frac{a}{a+w(b+c)}
 \label{ad_2}
\end{equation}

Some commonly used binary similarity measures are listed in Eq~\ref{ad_3} $\sim$ Eq~\ref{ad_7}, more measures please ref to surverys \cite{consonni2012new,todeschini2012similarity,choi2010survey}.
\begin{equation}
 S_{YULEQ}=\frac{ad-bc}{ad+bc}
\label{ad_3}
\end{equation}
\begin{equation}
 S_{TARANTULA}=\frac{a(c+d)}{c(a+b)}
\label{ad_4}
\end{equation}
\begin{equation}
 S_{MICHAEL}=\frac{4(ad-bc)}{(a+d)^2+(b+c)^2}
\label{ad_5}
\end{equation}
\begin{equation}
 S_{SIMPSON}=\frac{a}{min(a+b,a+c)}
\label{ad_6}
\end{equation}
\begin{equation}
 S_{BRAUN\&BANQUET}=\frac{a}{max(a+b,a+c)}
\label{ad_7}
\end{equation}

\section{Our Approach}
This section presents the GA-based searching for similarity measure in BNNs. First, we propose to represent the similarity measure as a combination of simple functions. Next, we introduce two prerequisites (encoding strategy and fitness function) in GA. Then, we describe GA-based searching in detail.

\subsection{Methodology}
In a typical binary convolution neural network, the dot product is adopted to measure the similarity between the inputs and weights. All elements in the inputs and weights are binary values: +1/-1, and encoded as 1/0. The costly floating-point dot product can be achieved by lightweight bit-wise operations, as shown in the Eq ~\ref{hh1}. $x$ is the full-precision the input, which is binarized as $x^b$. $W$ is the weight and binarized as $W^b$. XNOR is the bit-wise exclusive-nor operation, which perform the multiplication on a bit-wise level. Popcount is population-count operation, which perform the accumulation. 
\begin{equation}
  y = x \ast W \approx x^b \ast W^b = \text{popcount}(\text{XNOR}(x^b,W^b))
 \label{hh1}
\end{equation}

As discussed in Section II-C, there are four combinations of two binary variables: 1/1, 1/0, 0/1, 0/0, and $a$, $b$, $c$, $d$ represent the frequencies of the corresponding combinations. There are numerous binary similarity measures have been proposed for various tasks.
Most of them are functions of $a$, $b$, $c$, $d$.
We adopt this observation, assuming the similarity measure in BNNs  is function of $a$, $b$, $c$, $d$, as shown in Eq~\ref{hh2}. $a$, $b$, $c$, $d$ can be achieved by bit-wise operations AND, operasite ($\sim$) and bit count (popcount), As shown in Eq~\ref{hh3} -- Eq~\ref{hh6}. the bitwise inversion $\sim x^b$ and $\sim W^b$ can be reused.
\begin{equation}
  y = f(a,b,c,d)
 \label{hh2}
\end{equation}
\begin{equation}
  a = \text{popcount}(\text{AND}(x^b,W^b))
 \label{hh3}
\end{equation}
\begin{equation}
  b = \text{popcount}(\text{AND}(\sim x^b,W^b))
 \label{hh4}
\end{equation}
\begin{equation}
  c = \text{popcount}(\text{AND}(x^b,\sim W^b))
 \label{hh5}
\end{equation}
\begin{equation}
  d = \text{popcount}(\text{AND}(\sim x^b, \sim W^b))
 \label{hh6}
\end{equation}

Cross-correlation is typically used in  BNNs to measure similarity between inputs and weights,  which can be expressed as Eq ~\ref{hh7}. 
In this paper, we propose to discover more appropriate similarity measure for BNNs to improve the performance.
\begin{equation}
  y = a+d-(b+c)
 \label{hh7}
\end{equation}

\begin{figure*}[b]
\centering
\includegraphics[width=1.3\columnwidth]{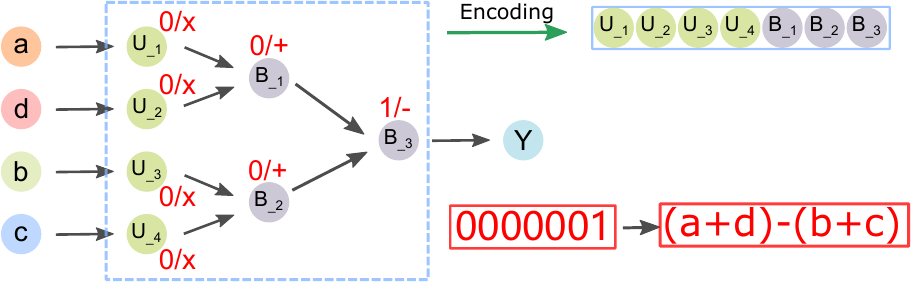}
\caption{Encoding.}
\label{fig:adbc}
\end{figure*}

\subsection{Encoding Strategy}

The similarity measure can not be too complicated, otherwise it will increase the difficulty and instability for network training. We propose using a graph with fixed structure to represent the similarity measure, each node stands for an operator.  There are 4 input nodes: $a$, $b$, $c$, $d$ in the graph. 
Considering $a$, $b$, $c$, and $d$ may be not equally important for the measure, four input nodes pass through independently
unary operator respectively, which can enlarge or reduce the input node adaptive.
Then the scaled matching frequencies $a$ and $d$, and the unmatched frequencies $b$ and $c$ are integrated with binary operators respectively. After another integration, the final output is generate.
As shown in Fig~\ref{fig:adbc}. $a$, $d$, $b$, $c$ are inputs, which are computed based on  Eq~\ref{hh3} -- Eq~\ref{hh6}, $Y$ is output.
$U$ stands for unary operator, that both input and output are one tensor with same size. $B$ stands for binary operator, which has two tensors as input and one tensor as output, all three tensor have the same size. The similarity measure consists of 4 unary operators and 3 binary operators, which can be encoded as a string with fixed length of 7: $U_1$ $U_2$ $U_3$ $U_4$ $B_1$ $B_2$ $B_3$. 
all operators are independent with each other.

A list of unary and binary operators are shown as follows:
\begin{itemize}
    \item \textbf{Unary operator:} $x$, $0$, $x^2$, $x^3$, $\sqrt{x}$, $log(x)$, $sin(x)$, $cos(x)$, $1.0/(1+e^{-x})$, $tan(x)$, $atan(x)$, $erf(x)$, $erfc(x)$, $e^{(-x)}$, $e^{-x^2}$, $\alpha$, $\alpha*x$, $\alpha+x$.
    \item \textbf{Binary operator:} $x+y$, $x-y$, $y-x$, $x*y$, $x/y$, $x/(x+y)$, $y/x$,  $y/(x+y)$, $max(x,y)$, $min(x,y)$, $x/(1+e^{-y})$, $y/(1+e^{-x})$,  $e^{(-|x-y|)}$, $e^{(-(x-y)^2)}$.
\end{itemize}

$0$ in unary operator means the output tensor is always $0$, which enables the pruning of redundant operators. $\alpha$ is learnable channel-wise parameter, which can be joint optimized with network parameters during back-propagation. 

There are 18 candidate functions for unary operator and 14 candidate operators for binary operator. In the encoding string, $U$ ranges from 0 to 17 and $B$ ranges from 0 to 13. As shown in Figure ~\ref{fig:adbc}, cross-correlation (Eq ~\ref{hh7}) can be encoded as 0000001. The former part 0000 represent the 4 unary operators are $x$, which is encoded as 0. The latter part 001 represent the 3 binary operators are $x+y$, $x+y$ and $x-y$. The search space is 18$\times$18$\times$18$\times$18$\times$14$\times$14$\times$14 = 288M.

\subsection{Fitness Function}

The fitness function in GA is used to evaluate the degree of goodness of candidate solution.
The fitness value indicates the quality and competitiveness of candidate individual in the population. The individual with better fitness value has higher chance to be selected for reproducing offspring.

In this paper, we focus on searching similarity measure for BNNs. The accuracy of classification on a reference dataset is adopted to evaluate the candidate individuals. As each trial assessment of the fitness evaluation requires the training of a BNN model, to be efficient, the small dataset Cifar10 is used to be the reference dataset. Not all well-known real-valued models can be seamlessly applied to BNNs, and given the widely adoption of ResNet structure in BNNs, we use the classical ResNet-18 model to perform the evaluation.  In particular, we train ResNet-18 with candidate individual on Cifar10 for 15 epochs, the top-1 validation accuracy is assigned to the candidate individual as fitness value.

\textbf{Early Rejection:} Since the candidate similarity measure is generated pseudo-randomly, some of them may be invalid (i.e. leading non-convergence during training). we adopt the early rejection strategy, the compute resources can be focusing on the promising candidates by ruling out invalid candidate individuals at very beginning. A rejection threshold $T$ is set. If the validation accuracy is less than $T$ after 1 epoch of training, the candidate individual will be discarded without further training. Otherwise, the training is continued up to 15 epochs.

The fitness function is as shown in Eq~\ref{fit5}, $T$ is the rejection threshold, $ACC_i$ stands for validation accuracy after training Resnet-18 (with candidate similarity measure) on Cifar10 for i epochs.
\begin{equation}
    Fitness(f) = \begin{cases}
    & ACC_{i = 1}   \qquad ACC_{i = 1} < T  \\
    & ACC_{i = 15}    \qquad \text{otherwise}        \\
    \end{cases}
    \label{fit5}
\end{equation}

\subsection{GA details}
The workflow of GA is shown in Algorithm~\ref{ga_flow}. Our GA starts with an initialization population with size $S$. Each individual in the population is evaluated and assigned with a fitness value. The population is ranked based on the fitness values from strongest to weakest. New offspring for the next generation is produced by three genetic operations: selection, crossover and mutation.

The new offspring is compared with the weakest (last) individual in the old population, as Line~27-30 in Algorithm~\ref{ga_flow}: \textbf{(i)} if the the new offspring has a higher fitness value, the last individual in the old population is deleted and the new offspring is insert in the population based on its fitness value. The next generation begins based on the new population. \textbf{(ii)} Otherwise, the next generation starts using the old population. During the evolution process, the size of  population keeps unchanged ($S$). Once the population has converged meaning that no more satisfactory solutions can be generated, the searching stops and the top-ranked individuals in the final population is the obtained promising solutions.

\begin{algorithm}[htb]
\caption{The workflow of genetic algorithm}
\label{ga_flow}
\KwIn {$S$: the size of population; $T$: the early reject threshold.}
\KwResult {Top-ranked individuals are promising solutions\;}
\tcp{Initialization:}
$i$ = 1; $POPU$ = []; $FITV$ = [] \;
\While {$i< S-1$}{
    randomly generate individual $M$ \;
    evaluate the fitness of $M$\;
    \If {fitness($M$) $> T$}{
            $POPU$.append($M$)\;
            $FITV$.append(fitness($M$))\;
            $i$++\;
    }
}
evaluate fitness of $M_0 = 0000001$\;
$POPU$.append($M_0$)\;
$FITV$.append(fitness($M_0$))\; 
Sort $POPU$ based on $FITV$ \;
\tcp{Evolution:}   
\While {Not Converage }{
    \tcp{Selection:}
    $n$ = randint(3)\;
    select $Parent_1$, $Parent_2$ based on $Selection_n$ \;
    \tcp{Crossover:}
    $k$ = randint(7) \;
    \eIf {randint(2)} 
      {$C1$ = $Parent_1[:k]$ + $Parent_2[k:]$;} 
      {$C1$ = $Parent_2[:k]$ + $Parent_1[k:]$;}
    \tcp{Mutation:}
    $m$ = randint(7) \;
    $C2$ = replace $C1[m]$ with a randomly gene\;
    \tcp{Compare:}
    Evaluate fitness of $C2$ \;
    \If{fitness($C2$) $> FITV[-1]$}
    {Delete $POPU[-1]$ and $FITV[-1]$ \;
    Insert $C2$ and fitness($C2$) in $POPU$ and $FITV$ based on fitness($C2$) \;}
}
\end{algorithm}

\vspace{5pt}\noindent\textbf{Initialization:}
Usually, there are two strategies for generating the initialization population: i) Random generation, which can ensure the diversity of the population; ii) Applying mutation on the existing sub-optimal solution, which is easy to convergence. We adopt a hybrid initiation strategy. First, 
by integrating $S-1$ randomly generate individuals (Line in A-1) with sub-optimal individual $0000001$ as initialization population (as Line~ 1-14 in Algorithm~\ref{ga_flow}). with early rejection, the invalid individuals is eliminated from population, So 

Generally, there are two strategies for generating the initialization population: Random generation or Applying mutation on the existing sub-optimal solution. We adopt a hybrid strategy. First, to ensure the diversity of population $S-1$ valid individuals are randomly generate (as Line~1-10 in Algorithm~\ref{ga_flow}). A individual $M$ is randomly generated and evaluated with fitness function. To ensure the valid of initialization population, the individual with a fitness value greater than rejection threshold $T$ will remain within the population.
Then sub-optimal individual $0000001$ and its fitness value are added to the population (as Line~10-13 in Algorithm~\ref{ga_flow}). The population is sorted in descending order based on the fitness values(as Line~14 in Algorithm~\ref{ga_flow}). The first individual has the highest fitness value.

\vspace{4pt}\noindent\textbf{Selection:} 
Selection is the process of selecting parent individuals, which will be used for reproducing new offspring for the next generation. Selection is crucial to the convergence rate of the GA as good parents drive individuals to a better and fitter solutions. To simulate the theory of "Survival of the Fittest", the individuals with better fitness have higher probabilities to be selected.
And for the diversity of population, each individual should have the chance to be selected. 

There are several well-known selection techniques, such as elitism, roulette wheel, rank, tournament, Boltzmann, stochastic universal sampling, etc. In our GA-based searching, two individuals are selected as parents. Considering the variability and efficiency of the algorithm, a hybrid selection strategy is adopted. As shown in Line~16-17 of Algorithm~\ref{ga_flow}, In each generation, we randomly choose one of the three candidate selection techniques: \emph{elitism selection}, \emph{tournament selection}, and \emph{proportionate selection}. Then two parent individuals are selected based on the picked technique.

\textbf{\emph{1) Elitism selection:}} The strongest individuals in the population are guaranteed to be selected. In our algorithm, the first two individuals in the ranked population are selected. Choosing the best individuals for reproduction could lead to a better chance of reproducing a better individual. However, it also leads to the individuals being close to each other and result in early convergence due to being trapped in a local maximum.

\textbf{\emph{2) Tournament selection:}} Subgroups of individuals are chosen from the larger population, and members of each subgroup compete against each other. Only one individual from each subgroup is chosen to reproduce. In our algorithm, the first individual is randomly selected from the population, and the second individual is randomly picked from the right part of the first one in the population. Thereby, the second individual has a lower fitness than the first one. Both the two individuals are chosen at random.

\textbf{\emph{3) Proportionate selection:}} Every individual can be selected with a probability which is proportional to its fitness. First,  fitness values of individuals are normalized by dividing the sum of all fitness values. Then normalized fitness values are accumulated, and  the accumulated fitness of the last individual should be 1.  Two random individuals are selected from the distribution by generating two random numbers between 0 and 1.

Elitism selection always chooses the best individuals whereas tournament selection and proportionate selection  stochastically choose individuals. The probabilistic combination of those three selection techniques can take the advantages of each technique, while avoiding their drawbacks, accelerating the convergence speed.

\vspace{4pt}\noindent\textbf{Crossover:} The crossover operator transmitted the best features of selected individuals to the next generation, which will have a better fitness value on average. The well-known crossover operators include \emph{single-point}, \emph{two-point}, \emph{k-point}, \emph{uniform}, \emph{partially matched}, \emph{ordered}, \emph{precedence preserving}, \emph{shuffle}, etc. The most common type is single point crossover, which is adopted in our algorithm.

In single-point crossover, the single crossover point is picked, then the offspring take one part from selected parent individuals. As shown in Line~18-23 of Algorithm~\ref{ga_flow}, the crossover point is randomly selected from 0 to 6. The two parent individuals are split into left part and right part depending on crossover point. The new offspring is formed by combining the left part of $Parent_1$ with the right part of $Parent_2$, or by combining the left part of $Parent_2$ with the right part of $Parent_1$.

\vspace{4pt}\noindent\textbf{Mutation:} During selection operator, the stronger individuals have higher chance to be selected. There is a tendency that the new offspring may become very similar after several generations, and the diversity of the population may decline, which could lead to population stagnation. Mutation is a mechanism to insert random genes in offspring to maintain the diversity of the population. The well-known mutation operators include \emph{displacement}, \emph{simple inversion}, and \emph{scramble mutation}. 

In our algorithm, we adopt single point mutation, which is one variant of displacement mutation, as illustrated in Line~24-25 of Algorithm~\ref{ga_flow}. First, a single mutation point is randomly picked. Then the selected gene is replaced by a new randomly generated gene. 
In encoding string, all unary operators stay on the left and binary operators stay on the right. Based on the mutation point, we figure out whether the mutation refers to operator $U$ or $B$. If $U$, a replacement gene is randomly generated from the 18 unary operators; otherwise the replacement gene is randomly generated from the 14 binary operators.

\section{Experiment}
First, we perform the GA-based searching, then the promising similarity measures are listed and evaluated on Cifar10 and Cifar100 with several network models. We use Pytorch to do the searching and evaluation.

\subsection{Experiment Setting}

\textbf{BNN model:} Usually in BNNs, the first and the last layer are full-precision. Using full-precision for the first layer is to conserve the maximum information flow from the input images. Using the full-precision for the last layer is to conserve the maximum state-space before the final output. The first and last layer of BNN models used in this paper always retain in full-precision, while other layers are binary.

\textbf{GA setting}: The size of population $S$ is set to be 30.

During evaluation of fitness value, the ResNet-18 with candidate similarity measure is trained on Cifar10 for 15 epoches, and the top-1 validation accuracy is assigned as fitness value. During training, the standard pre-processing is adopt, the batch size is 128. The Adam optimizer with $betas=(0.9,0.999)$ is  adopted with a constant learning rate 5e-3.

The early rejection threshold $T$ is empirically set as 11\% in our fitness evaluation. With the early rejection strategy, the compute resources can be allocated focusing on the most promising candidates. To further enhance searching efficiency, the threshold can be slowly increased as the search continues, so that it can early reject more invalid candidate individuals and preserve promising solutions. This is based on the intuition that as time goes on, the descendants of the GA tend to be more well-behaved than in the early stage of the search. Increasing the early rejection threshold thus improves the search efficiency. As the searching goes on, the threshold $T$ increases to be 25\%, 35\%, 40\%.

\subsection{Similarity Measures Obtained by GA-based Search}
The promising similarity measures obtained by GA-based search are list in Table~\ref{M_fun}. The first line is the widely used cross-correlation which is the default measure in both DNNs and BNNs. we use the cross-correlation as baseline. The M1 $\sim$ M10 is obtained from GA-based search. The row 2 $\sim$ 5 are the unary operators, that correspond to $U_1$ $\sim$ $U_4$ in Figure~\ref{fig:adbc}, with $a$, $d$, $b$, $c$ as input and $a'$, $d'$, $b'$, $c'$ as output. The last row is the binary operators, as $B_1$ $\sim$ $B_3$ in Figure~\ref{fig:adbc}, with $a'$, $d'$, $b'$, $c'$ as the inputs.

\begin{table}[htbp]
\centering
\caption{Promising Similarity Measures}
\begin{tabular}{c | c| c| c |c |c}
\hline
\hline
\textbf{Measure}  & \textbf{ $a'(a)$ }& \textbf{ $d'(d)$} & \textbf{ $b'(b)$ }& \textbf{ $c'(c)$} & \textbf{ $f(a'd'b'c')$} \\
\hline
\hline
 baseline    & $a$ &  $d$ & $b$ &  $c$ & $(a'+d')-(b'+c')$  \\
\hline
M1 & $a^3$ &  $d$ & $b^3$ &  $c$ & $\frac{b'-c'}{a'+d'}$  \\ 
\hline
M2 & $a^3$ &  $d$ & $b^3$ &  $sigmoid(c)$ & $\frac{b'+c'}{a'+d'}$  \\ 
\hline
M3 & $a^3$ &  $d$ & $b^3$ &  $c$ & $\frac{b'+c'}{a'+d'}$  \\ 
\hline
M4 & $a^3$ &  $d$ & $b^3$ &  $ sin(c)$ & $\frac{b'-c'}{a'+d'}$  \\ 
\hline
M5 & $a^3$ &  $e^{-d*d}$ & $b$ &  $erf(c)$ & $\frac{b'-c'}{a'+d'}$  \\  
\hline
M6 & $a^3$ &  $d$ & $b^3$ &  $c$ & $\frac{b'*sigmoid(c')}{a'+d'}$ \\
\hline
M7 & $a^3$ &  $\alpha$ & $b^3$ &  $c$ & $\frac{a'+d'}{b'+c'}$  \\
\hline
M8 & $a^3$ &  $d$ & $b^3$ &  $e^{-c}$ & $\frac{b'+c'}{a'+d'}$  \\ 
\hline
M9 & $a^3$ &  $d^2$ & $b^3$ &  $atan(c)$ & $\frac{b'}{c'*(a'+d')}$  \\ 
\hline
M10 & $a^3$ &  $d^2$ & $b^3$ &  $sigmoid(c)$ & $\frac{b'+c'}{a'+d'}$  \\ 
\hline
\hline
\end{tabular}

\label{M_fun}
\end{table}

There are some observations:
\begin{itemize}
    \item For the similarity measure in BNNs, the inputs are not equally important. $a$, $b$, $d$, $c$ are the ranked queue from most significant important to less important.
    The frequencies of positive matched (attributes both inputs share) $a$ is the most important, unary operators in the 10 promising measures are always $x^3$, which is the operation with the largest magnification in unary operations. 
    Except for M5, the unary operators of $b$ are $x^3$. Most of the unary operators of $d$ are $x$ (unchanged), while the others enlarge input value (M9 and M10), shrink input value (M5) or is learned constant value (M7). $c$ has the weakest impact on the measure, as most of  unary operators shrink the value of input.
    \item For binary operation, $B_1$ (the integration of $a'$ and $d'$) is addition operator. $B_2$ (the integration of $b'$ and $c'$) is addition (mostly), subtraction (M1, M4 and M5) or division (M6 and M9). $B_3$ are always division operator. 
    It can be seen that comparing to other operators, the ratio scheme is a good way to measure the binary similarity for BNNs. 
\end{itemize}

\subsection{Evaluation}
The image classification is commonly used to test the performance of deep neural network, so is the BNNs. To verify the performance and generalization of similarity measures, we evaluate the measures on different datasets with different network models. In this section, we evaluate the measures listed in Table~\ref{M_fun} on Cifar10 and Cifar100 with ResNet-18, ResNet-34, NIN-Net and VGG13.

For the fairness of evaluation, all the hyper-parameters are same except for the similarity measure. The standard data preprocessing strategy is adopted, the batch size is 128. We train all network models for 300 epochs and Adam optimizer with $betas=(0.9, 0.999)$ is adopted for optimization. The initial learning rate is 5e-3, and is divided by 5 at the 80th, 150th, 200th, 240th, and 270th epoch.

\vspace{5pt}\noindent\textbf{Cifar10:}
We evaluate the similarity measures on Cifar10, the top-1 validation accuracy are shown in Table~\ref{M_cifar10}. The measure of first line is cross-correlation, which is the baseline. M1 $\sim$ M10 are obtained from GA-based search as list in Table~\ref{M_fun} .

\begin{table}[h]
\centering
\caption{Evaluation of Similarity Measures on Cifar10}
\label{M_cifar10}
\begin{tabular}{c| c| c |c c} 
\hline
\hline
\textbf{Measure} & \textbf{ResNet-18}  & \textbf{ResNet-34}  & \textbf{NIN-Net}  & \textbf{VGG13}  \\
\hline
\hline
(a+d)-(b+c)& 90.51  & 91.00   &  86.46   &  88.83  \\
\hline
M1 & 91.00  &  91.59  &  86.63   &  88.70  \\
\hline
M2 & 90.98  & 91.56   &  86.54   &  89.20  \\
\hline
M3 &  90.83 &  91.50  &  86.60   &  89.32  \\
\hline
M4 & \textbf{91.11}  &  91.65  &  86.44   &  89.17  \\
\hline
M5 &  90.84 &  90.88  &  86.26   &  88.62  \\
\hline
M6 & \textbf{91.14}  &  91.37  &  86.48   &  89.09  \\
\hline
M7 & 91.07  &  91.31  &  \textbf{86.83}   &  88.96  \\
\hline
M8 & 91.00  &  \textbf{91.75}  &  86.19   &  \textbf{89.34 } \\
\hline
M9 & 90.81  &  \textbf{91.66}  &  \textbf{86.79}   &  \textbf{89.38 } \\
\hline
M10 & 90.87  &  91.62  &  86.51   &  89.24  \\
\hline
\hline
\end{tabular}
\end{table}

For ResNet18, all the 10 obtained measures improve the performance compared to the baseline. The top 2 measures are M6 and M4, which improve accuracy by 0.63\% and 0.6\%, respectively.
For ResNet34, the top measures are M8 and M9, which has performance improvement of 0.75\% and 0.66\%, respectively. For NIN-Net, the top measures are M7 and M9, which have 0.37\% and 0.33\% performance improvement respectively. For VGG13, M9 and M8 are the top measures, which improve the accuracy by 0.55\% and 0.51\% respectively.

For the 10 obtained measures, M2, M3, M6, M7, M9 and M10 outperform the baseline on Cifar10, while the most improvement is attain by M8 on ResNet-34 with 0.75\% accuracy improvement.

\vspace{5pt}\noindent\textbf{Cifar100:}
We evaluate the similarity measures which list in Table~\ref{M_fun} on Cifar100, the top-1 and top-5 validation accuracy are shown in Table~\ref{M_cifar100}.
\begin{table}[!h]
\centering
\caption{Evaluation of Similarity Measures on Cifar100}
\label{M_cifar100}
\begin{tabular}{c | c| c| c} 
\hline
\hline
\textbf{Measure} & \textbf{ResNet-34(Top-1/5)} & \textbf{NIN-Net(Top-1/5)} & \textbf{VGG13 (Top-1/5)}\\
\hline
\hline
(a+d)-(b+c)   &  62.04/86.08  &  56.63/83.77   &  55.56/81.06  \\ 
\hline
M1  &  63.85/87.25  &  \textbf{57.77/85.16}   &  \textbf{58.42/83.39 } \\ 
\hline
M2  &  63.22/86.73  &  56.61/84.07   &  58.04/84.07  \\ 
\hline
M3 & 62.78/86.58   &  56.78/84.18   &  57.63/83.07  \\ 
\hline
M4 &  63.65/86.99  &  57.48/84.90   &  58.26/83.58  \\ 
\hline
M5  &  61.49/85.26  &  57.33/84.17   &  \textbf{58.95/84.44 }  \\ 
\hline
M6  &   63.06/86.87 &  56.72/84.61   &  57.56/82.63  \\ 
\hline
M7 &  63.64/87.05  &  \textbf{57.84/85.20}   &  58.06/83.90  \\
\hline
M8 &  63.42/87.19  &   57.15/84.56  &  57.97/83.46  \\ 
\hline
M9 &  \textbf{64.96/88.07 } &  56.29/83.44   &  57.59/83.22  \\ 
\hline
M10  & \textbf{64.56/87.66}   &  57.09/84.71   &  57.14/82.39  \\ 
\hline
\hline
\end{tabular}
\end{table}

For ResNet-34, the top improvement of similarity measure are M9 and M10, which have 2.92\% and 2.52\% accuracy improvement compared to baseline. For NIN-Net, the top measures are M7 and M1, which achieve 1.21\% and 1.14\% accuracy improvement respectively. For VGG13, M5 and M1 improve the accuracy by 3.39\% and 2.86\%, which achieve the top performance.  

Except for M2 and M9, other measures improve the accuracy on Cifar100 of the three network models. M1, M4, and M7 have relatively better improvement.

\vspace{5pt}

Overall, the similarity measures obtained by GA-based search improve the performance of BNNs. The accuracy improvement on Cifar10 is minor, as the accuracy on Cifar10 is already very high (over 90\%), and there is little room for improvement. The accuracy on Cifar100 is significantly improved, up to 3.39\%.

\section{Conclusion}
BNNs suffer severe accuracy degradation, which limit its applicability in various domains. In this paper, we focus on identifying BNN-tailored similarity measure to improve performance. We represent the similarity measure with a fixed graph, which can be encoded as a string with fixed length. Several promising similarity measures are obtained through GA-based searching. We evaluate the obtained similarity measures on Cifar10 and Cifar100 with several network models.
Results shown that the newly discovered similarity measures can improve the accuracy of BNNs, which up to 3.39\% accuracy improvement on Cifar100.

\ifCLASSOPTIONcaptionsoff
  \newpage
\fi



\bibliographystyle{IEEEtran}
\bibliography{ref}
%

\end{document}